\documentclass[10pt,twocolumn,letterpaper]{article}

\usepackage{3dv}
\usepackage{times}
\usepackage{epsfig}
\usepackage{graphicx}
\usepackage{amsmath}
\usepackage{amssymb}
\usepackage{fancychapters}
\usepackage[autostyle]{csquotes}  

\usepackage{balance}

\usepackage[pagebackref=true,breaklinks=true,letterpaper=true,colorlinks,bookmarks=false]{hyperref}

\usepackage{siunitx}
\usepackage{subfigure}
\usepackage{diagbox}
\usepackage{multirow}

\def\mean{\mathop{\operator@font mean}}

\usepackage[nocompress]{cite}

\def\ie{\emph{i.e.}}

\threedvfinalcopy 

\ifthreedvfinal\fi
\begin{document}

\title{Combining Deep and Depth: \\
Deep Learning and Face Depth Maps for Driver Attention Monitoring}

\author{Guido Borghi\\
Department of Engineering ``Enzo Ferrari'' - AImageLab\\
University of Modena and Reggio Emilia\\
\{name.surname\}@unimore.it}

\maketitle

\begin{abstract}
Recently, deep learning approaches have achieved promising results in various fields of computer vision. 
In this paper, we investigate the combination of deep learning based methods and depth maps as input images to tackle the problem of driver attention monitoring.
Moreover, we assume the concept of attention as Head Pose Estimation and Facial Landmark Detection tasks. 
Differently from other proposals in the literature, the proposed systems are able to work directly and based only on raw depth data. All presented methods are trained and tested on two new public datasets, namely Pandora and MotorMark, achieving state-of-art results and running with real time performance. 
\end{abstract}

\section{Introduction}
The introduction of semi-autonomous and autonomous driving vehicles and their coexistence with traditional cars is going to increase the already high interest about driver attention studies. 
Very likely, automatic pilots and human drivers will share the control of the vehicles, and the first will need to call back the latter when needed:  in this case, the monitoring of the driver attention level is a key-enabling factor. 
In addition, legal implications will be raised~\cite{Rahman4021}.\\
In this context, the human face is one of the richest source of information to monitor the \textit{visual} driver distraction~\cite{craye2015driver}, \textit{i.e.} when the driver's eyes are not looking at the road. \\
In this paper, we assume the concept of attention as a Head Pose Estimation problem, in conjunction with an analysis about the state of salient elements belonging to the face (\textit{i.e.} eyes, mouth).
These two tasks are exploited, for instance, to infer the coarse gaze~\cite{pradhan2012measurement} or to detect the driver drowsiness through the state of mouth or eyes~\cite{benfold2011unsupervised, abtahi2011driver}.\\
Moreover, only depth images, acquired with near-infrared sensors, are used as input data in order to develop methods that work even during the night or with bad light source conditions. 
Specifically, we investigate the potentiality of depth images combined with deep learning based methods, a research topic not fully exploited, since the release of cheap but accurate small-sized 3D sensors brings up new opportunities in this field and much high-quality depth maps. 

Summarizing, we present a study about algorithms that satisfy automotive requirements (light invariance, real time performance, occlusion reliability), focusing on Head Pose Estimation and Facial Landmark Detection tasks, based only on depth images.

\begin{figure}[t!]
    \centering
    \includegraphics[width=0.8\columnwidth]{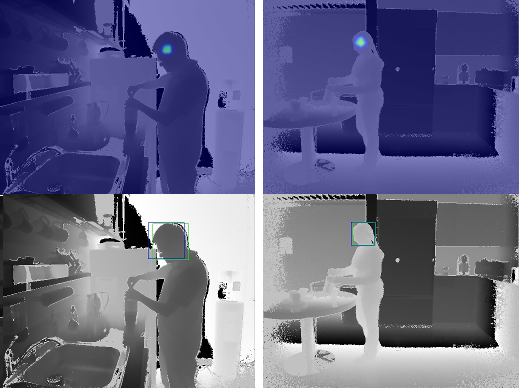}
    \caption{Head Detection task: depth frames with the predicted probability maps (first row) and depth maps with head predictions (green) and ground truth annotations (blue) in the second row. }
    \label{fig:head_detection}
\end{figure}

\section{Head Detection}
Head detection is the ability to detect and localize one or more heads in a given input image~\cite{hjelmaas2001face} and is a key element for applications based on the analysis of the head.\\
Most of the current research approaches are based on images taken by conventional visible-light cameras -- \textit{i.e.} RGB or intensity cameras -- and only few works tackle the problem of head detection in \textit{depth images}.\\
Head detection methods based on depth images have several advantages over methods based on 2D information that generally suffer the complexity of the background and when subject's head has not a consistent color or texture. \\
In \cite{diego2018fully}, we propose a \textit{Fully Convolutional Network} that is able to output a probability map based on head locations, given a depth input map. 
Experimental results show the good accuracy, the reliability and the speed performance (more than 30 fps) of the framework. 
This method is currently the state-of-art for head detection in the wild with only depth images. Sample output images are reported in Figure \ref{fig:head_detection}.\\
Furthermore, we investigate a CNN used as a binary classifier, designed to classify candidate patches as head or non-head \cite{ballotta2017head}: in this case, depth maps are exploited to deal with the scale of the target object.

\begin{figure}[t!]
    \centering
    \includegraphics[width=0.9\columnwidth]{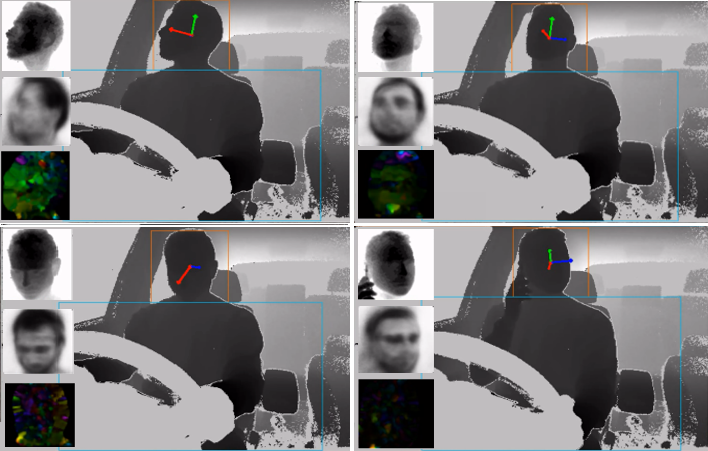}
    \caption{Head Pose Estimation task: samples from \textit{POSEidon} framework~\cite{borghi2017poseidon} output. Depth maps, \textit{Face-from-Depth} and \textit{Motion Images} are depicted on the left of each frame.}
    \label{fig:poseidon}
\end{figure}

\section{Head Pose Estimation}
Head Pose Estimation is the ability to infer the orientation of a person's head relating to the view of the acquisition device~\cite{murphy2009head}. The head pose in a 3D space can be expressed as a triplet ($\phi, \theta, \rho$) that corresponds to three angles: \textit{yaw}, \textit{pitch} and \textit{roll}.
Starting from head localization, we propose a framework, called \textit{POSEidon}~\cite{borghi2017poseidon}, to estimate head and shoulder poses, measured as continuous rotation angles. 
To this aim, a new triple regressive CNN architecture is proposed, that combines raw depth maps, \textit{Motion Images} (computed with the \textit{Farneback}~\cite{farneback2003two} algorithm) and generated faces from the corresponding depth images. \\
One of the most innovative contribution is a \textit{Face-from-Depth} network, that is able to reconstruct gray-level faces directly from depth images. This solution derives from the awareness that intensity face images are very useful to detect head pose~\cite{ahn2014real}: we would like to have similar benefits, without having intensity data. \\
\textit{POSEidon} is the current state-of-art in the head pose estimation task with depth maps. Sample outputs are depicted in Figure \ref{fig:poseidon}.
We investigate other several approaches \cite{venturelli2016deep}, including \textit{Siamese} networks \cite{venturelli2016depth} and an embedded implementation \cite{borghi2017embedded} for \textit{plug-in} and \textit{easy-to-install} car systems.

\section{Facial Landmark Detection}
\begin{figure}[b!]
    \centering
    \includegraphics[width=0.6\columnwidth]{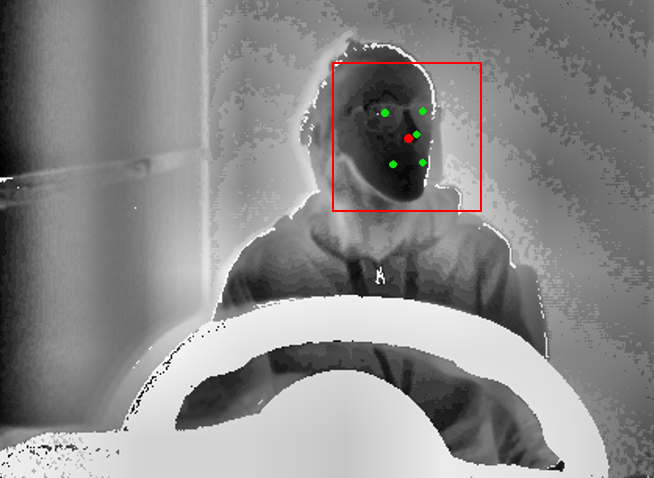}
    \caption{Facial Landmark Localization task: landmarks are reported with green circles and the square crop for face extraction is depicted with a red rectangle. }
    \label{fig:landmarks}
\end{figure}

A reliable localization of facial landmarks -- \textit{i.e.}, the ability to infer the position of prominent face elements relative to the view of the acquisition device -- is one of the basic components to conduct driver physical state investigation, through eyes or mouth direct monitoring~\cite{reddy2014driver} and facial expressions recognition~\cite{tie2013automatic}, as reported in literature.\\
We propose a deep-based approach specifically designed for real time facial landmarks localization in the automotive context, through a regression manner approach \cite{frigieri2017fast}.
The presented method is one of the few works that combines depth images and a deep architecture to localize facial landmarks in a regression manner: this explains the lack of database and competitors.\\
The model architecture is designed to deal with two main issues: low memory requirements and real time performance.
The proposed method relies only on depth data to achieve a good reliability in presence of illumination changes. A visual sample is reported in Figure \ref{fig:landmarks}.

\section{Face Generation}
As mentioned before, we also investigate the face generation task, inspired by the \textit{Privileged Information} approach~\cite{vapnik2009new}, in which the main idea is to add knowledge at training time -- the generated faces -- in order to improve the performance of the presented systems at testing time. Experimental results confirm the effectiveness of this approach.

\subsection{Face-from-Depth}
\textit{Is it possible to generate gray-level face images from the corresponding depth ones?} \\
\textit{Face-from-Depth} architecture~\cite{borghi2017face, borghi2017poseidon, fabbri2018domain} is designed to tackle this task and is one of the most innovative elements of \textit{POSEidon} framework. 
Due to illumination issues, the appearance of the face is not always available in many scenarios, \textit{e.g.} inside a vehicle. On the contrary, depth maps are invariant to illumination conditions but lack of texture details.
Moreover, we adopt some vision tasks as \textit{Perceptual Probes} for performance evaluation. 
We assess that the face-to-face translation is acceptable if new generated faces (from depth input) exhibit similar proprieties of original faces, \textit{i.e.} categorical attributes are maintained across domains.

\begin{figure}[t!]
    \centering
    \includegraphics[width=0.7\columnwidth]{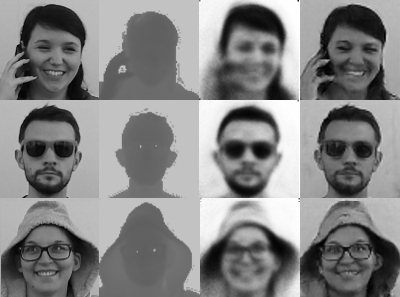}
    \caption{Gray-level images (first column) and the corresponding depth faces (second column); then, face images taken from \cite{borghi2017poseidon} and finally the output of the \textit{Face-from-Depth} network~\cite{borghi2017face}. }
    \label{fig:face_from_depth}
\end{figure}

\subsection{Depth-from-Face}
\textit{Is it possible to generate depth face maps from the corresponding gray-level ones?} \\
Depth estimation is a task at which humans naturally excel thanks to the presence of two high-quality stereo cameras (\ie~the human eyes) and an exceptional learning tool (\ie~the human brain). 
Even though depth estimation is a natural human brain activity, the task is an ill-posed problem in the computer vision context, since the same 2D image may be generated by different 3D maps. \\
In \cite{pini2018learning}, an \textit{adversarial} approach~\cite{goodfellow2014generative} is employed to effectively train a \textit{Fully Convolutional Autoencoder} that is able to estimate facial depth maps from the corresponding gray-level images. 
To the best of our knowledge, this is one of the first attempts to tackle this task through an adversarial approach that, differently from global depth scene estimation, involves small sized objects, full of details: the human faces.
Furthermore, we show that the model is capable of predicting distinctive facial details by testing the generated depth maps through a deep model trained on authentic depth maps for the Face Verification task.

\begin{figure}[b!]
    \centering
    \includegraphics[width=0.7\columnwidth]{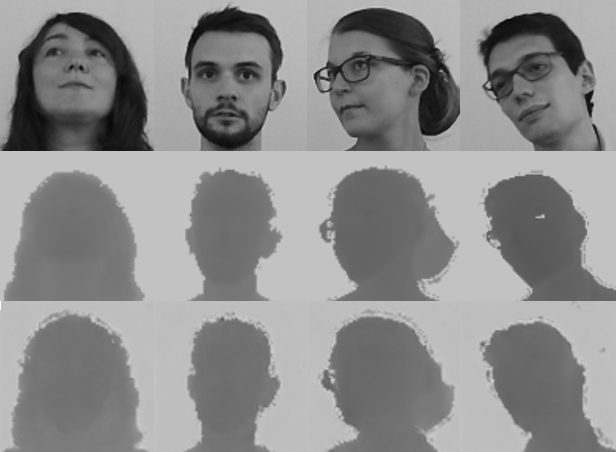}
    \caption{Sample outputs of the \textit{Depth-from-Face} network~\cite{pini2018learning} (last row). In the first and second rows, original gray-level and depth maps are reported, respectively.}
    \label{fig:depth_from_face}
\end{figure}

\section{New datasets collected}
All the previous methods are trained and tested thanks to two recent datasets.
Due to the lack of datasets containing face depth data, we decide to acquire two high-quality datasets, reproducing the automotive context.

\subsection{Pandora dataset}
We collect a new challenging dataset, called \textit{Pandora}~\cite{borghi2017face}, specifically created for the Head Pose Detection task. 
A frontal fixed device (\textit{Microsoft Kinect One}) acquires the upper body part of the subjects, simulating the point of view of a camera placed inside a car dashboard. 
\textit{Pandora} contains 110 annotated sequences using 10 male and 12 female actors. Each subject has been recorded five times. 
The dataset contains more than 250k full resolution RGB ($1920 \times 1080$) and depth ($512 \times 424$) images.  \\
\textit{Pandora} is the first publicly available dataset which contains \textit{shoulder angles}, in addition to the head pose annotations and \textit{wide angle ranges}. Moreover, it contains also \textit{challenging camouflage}: garments as well as various objects are worn or used by the subjects to create head and/or shoulder occlusions.

\subsection{MotorMark dataset}
MotorMark dataset~\cite{frigieri2017fast} includes both RGB and depth images, annotated with facial landmark coordinates. 
Frames are acquired through a \textit{Microsoft Kinect One}. It is composed by more than 30k frames, with a resolution of $1280 \times 720$ and $515 \times 424$, respectively.
Subjects (35 subjects in total) are standing in a real car dashboard and performs real inside-car actions.\\
The annotation of 68 landmark positions on both RGB and depth frames is available, following the \textit{ISO MPEG-4} standard. The ground truth has been manually generated. The user was provided with an initial estimation provided by the \textit{dLib} libraries, which gives landmark positions on RGB images. The projection of the landmark coordinates on the depth images is carried out exploiting the internal calibration tool of the \textit{Microsoft Kinect SDK}.

\section{Conclusion and Future Work}
In this paper, we investigate the combined use of deep learning methods and depth maps to monitor the driver attention, through the Head Pose Estimation and the Facial Landmark Detection tasks.
Experimental results confirm the potentiality and the feasibility of the presented methods. 
We plan to investigate the Face Verification task based only depth data as input.

\balance

{\small
\bibliographystyle{ieee}
\bibliography{egbib}
}

\end{document}